# OVERSAMPLING LOG MESSAGES USING A SEQUENCE GENERATIVE ADVERSARIAL NETWORK FOR ANOMALY DETECTION AND CLASSIFICATION

Amir Farzad[1] and T. Aaron Gulliver[2]

[1]Department of Electrical and Computer Engineering, University of Victoria, PO Box 1700, STN CSC, Victoria, BC Canada V8W 2Y2
ORCiD: 0000-0003-2499-7696
amirfarzad@uvic.ca

[2]Department of Electrical and Computer Engineering, University of Victoria, PO Box 1700, STN CSC, Victoria, BC Canada V8W 2Y2
ORCiD: 0000-0001-9919-0323
agullive@ece.uvic.ca

## ABSTRACT

*Dealing with imbalanced data is one of the main challenges in machine/deep learning algorithms for classification. This issue is more important with log message data as it is typically very imbalanced and negative logs are rare. In this paper, a model is proposed to generate text log messages using a SeqGAN network. Then features are extracted using an Autoencoder and anomaly detection is done using a GRU network. The proposed model is evaluated with two imbalanced log data sets, namely BGL and Openstack. Results are presented which show that oversampling and balancing data increases the accuracy of anomaly detection and classification.*

## 1. INTRODUCTION

Logs are commonly used in software systems such as cloud servers to record events. Generally, these unstructured text messages are imbalanced because most logs indicate that the system is working properly and only a small portion indicate a significant problem. Data distribution with a very unequal number of samples for each label is called imbalanced. The problem of imbalanced data has been considered in tasks such as text mining [1], face recognition [2] and software defect prediction [3].

The imbalanced nature of log messages is one of the challenges for classification using deep learning. In binary classification, there are only two labels, and with imbalanced data, most are normal (denoted major) logs. The small number of abnormal (denoted minor) logs makes classification difficult and can lead to poor accuracy with deep learning algorithms. This is because the normal logs dominate the abnormal logs. Oversampling and undersampling are two methods that can be used to address this problem. In undersampling, the major label samples are reduced so the number is similar to the minor label samples. A serious drawback of undersampling is loss of information [4]. In oversampling, the number of minor label samples is increased so it is similar to the number of major label samples. Recently, a generative adversarial network (GAN) [5] was proposed for generating images and showed good results in generating data which is similar to actual data such as with image captions [6]. GANs are able to generate more abstract and varied data than other algorithms [7].

In this paper we propose a model to deal with imbalanced log data by oversampling text log messages using a Sequence Generative Adversarial Network (SeqGAN) [8]. The resulting data is then used for anomaly detection and classification with Autoencoder [9] and Gated Recurrent Unit (GRU) [10] networks. An Autoencoder is a feed-forward network that has been shown to be useful for extracting important information from data. Autoencoders have been applied to many tasks such as probabilistic and generative modeling [11] and representation learning [12]. A GRU is a Recurrent Neural Network (RNN) which has been employed in tasks such as sentiment analysis [13] and speech recognition [14]. The proposed model is evaluated using two labeled log message data sets, namely BlueGene/L (BGL) and Openstack. Results are presented which show that the proposed model with oversampling provides better results than the model without oversampling.

The main contributions of this paper are as follows.

1. A model is proposed for log message oversampling for anomaly detection and classification.
2. The proposed model is evaluated using two well-known data sets and the results with and without oversampling are compared.

The rest of the paper is organized as follows. In Section 2 the Autoencoder, GRU and SeqGAN architectures are presented and the proposed model is described. The experimental results and discussion are given in Section 3. Finally, Section 4 provides some concluding remarks.

## 2. SYSTEM MODEL

In this section, the Autoencoder, GRU and SeqGAN architectures employed are given along with the proposed network model.

### 2.1. AUTOENCODER ARCHITECTURE

An Autoencoder is a feed-forward multi-layer neural network with the same number of input and output neurons. It is used to learn a more efficient representation of data while minimizing the corresponding error. An Autoencoder with more than one hidden layer is called a deep Autoencoder [15]. A reduced dimension data representation is produced using encoder and decoder hidden layers in the Autoencoder architecture. Backpropagation is used for training to reduce the loss based on a loss function. Figure 1 shows the Autoencoder architecture with an input layer, two hidden layers, and an output layer.

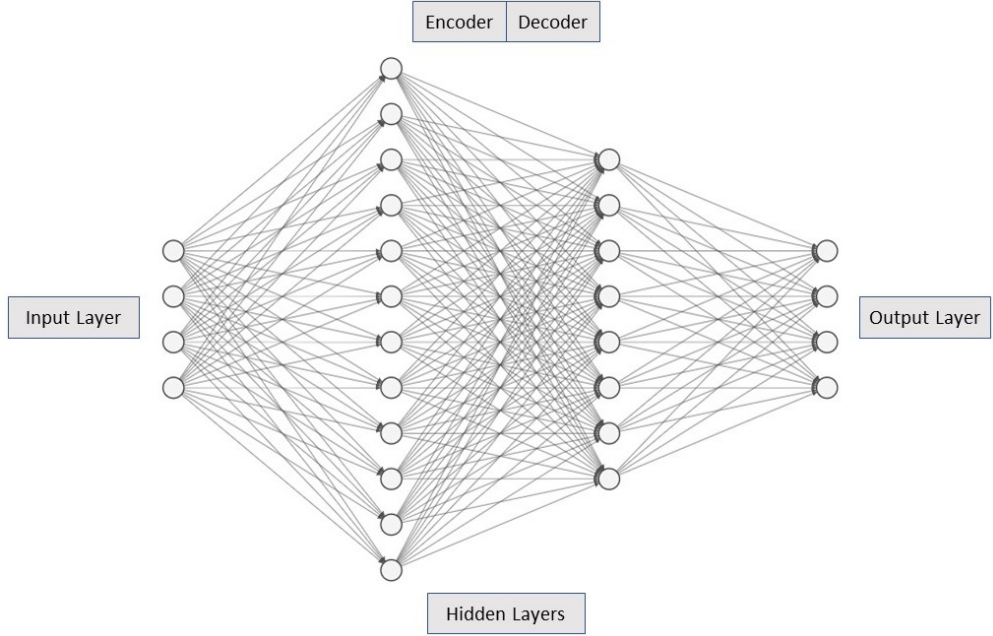

Figure 1. Autoencoder architecture with an input layer, two hidden layers, and an output layer.

## 2.2. GRU Architecture

A Gated Recurrent Unit (GRU) is a type of RNN network which is a modified version of an LSTM network [16]. It has a reset gate and an update gate. The reset gate determines how much information in a block should be forgotten and is given by

$$r_t = \sigma(W_r x_t + U_r h_{t-1} + b_r), \tag{1}$$

where $b_r$ is the bias vector, $\sigma$ is the sigmoid activation function and $W_r$ and $U_r$ are the weight matrices. The update gate decides how much information should be updated and can be expressed as

$$z_t = \sigma(W_z x_t + U_z h_{t-1} + b_z), \tag{2}$$

where $W_z$ and $U_z$ are the weight matrices and $b_z$ is the bias vector. The block output at time $t$ is

$$h_t = z_t \odot h_{t-1} + (1 - z_t) \odot \tanh(W_h x_t + U_h (r_t \odot h_{t-1}) + b_h), \tag{3}$$

where $b_h$ is the bias vector and $W_h$ and $U_h$ are the weight matrices. A GRU block is shown in Figure 2.

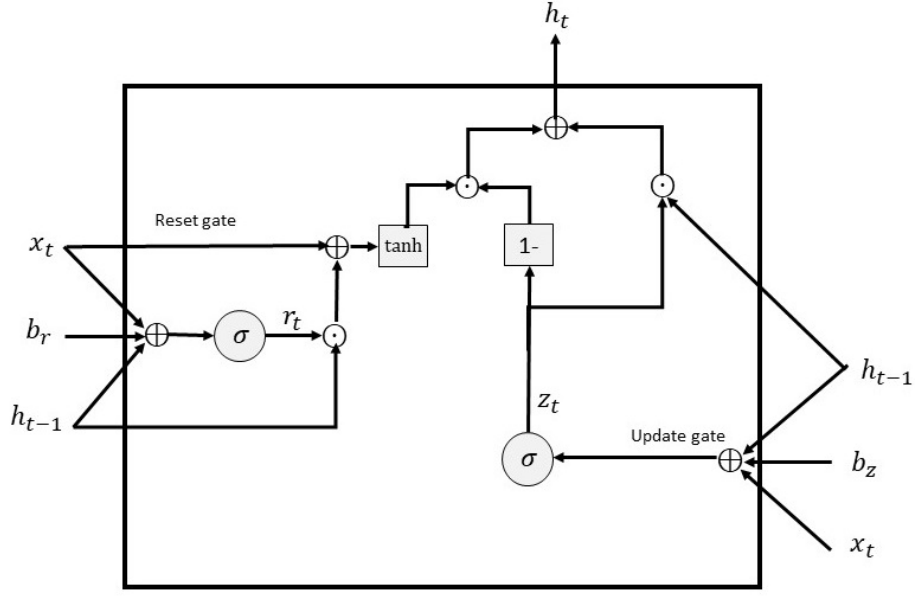

Figure 2. A GRU block with reset gate, update gate, and tangent hyperbolic and sigmoid activation functions.

### 2.3. SEQGAN ARCHITECTURE

A SeqGAN consists of a Generator ($G$) and a Discriminator ($D$). The Discriminator is trained to discriminate between real data (sentences) and generated sentences. The Generator is trained using the Discriminator using the reward function with policy gradient [17]. In SeqGAN, the reward for a sentence is computed and the Generator is regulated using the reward with reinforcement learning. Generator $G_\theta$ is trained with a real data set to produce a sentence

$$Y_{1:T} = \{y_1, \ldots, y_t, \ldots, y_T\}, y_t \in \mathcal{Y},$$

where $\mathcal{Y}$ is the vocabulary of candidate words. This should produce a sentence that is close to real data. This is a reinforcement learning problem which considers $G_\theta$ to produce an action $a$ (next word $y_t$) given the state $s$ (previously generated words $Y_{1:t-1}$). SeqGAN trains the Discriminator $D_\phi$ as well as the Generator $G_\theta$. $D_\phi$ is trained to discriminate between real data and data generated from $G_\theta$. Words are generated by $G_\theta$ each time step but $D_\phi$ only computes rewards for full sentences. Hence, the rewards for intermediate states are estimated using Monte Carlo (MC) search and are given by

$$Q_{D_\phi}^{G_\theta} = (s = Y_{1:t-1}, a = y_t) =$$

$$\begin{cases} \frac{1}{N} \sum_{n=1}^{N} D_\phi(Y_{1:T}^n), Y_{1:T}^n \in MC(Y_{1:t}; N) & \text{if}: t < T, \\ D_\phi(Y_{1:t}) & \text{if}: t = T, \end{cases} \qquad (4)$$

where $Q_{D_\phi}^{G_\theta}$ is the action-value function which is the expected reward from the Discriminator, $T$ is the sentence length and $N$ is the number of the sentences in the MC search, $Y_{1:T}^n$ is the $n$th sentence in the MC search, and $D_\phi(Y_{1:T}^n)$ is the probability of the $n$th sentence being denoted real by the Discriminator. After the reward is computed, the Generator $G_\theta$ is updated via the policy gradient which is the gradient of the objective function and is given by

$$\nabla_\theta J(\theta) \simeq \frac{1}{T}\sum_{t=1}^{T}\sum_{y_t \in \mathcal{Y}} \nabla_\theta G_\theta(y_t | Y_{1:t-1}) Q_{D_\phi}^{G_\theta}(Y_{1:t-1}, y_t)$$
$$= \frac{1}{T}\sum_{t=1}^{T} \mathbb{E}_{y_t \sim G_\theta(y_t|Y_{1:t-1})}[\nabla_\theta \log G_\theta(y_t | Y_{1:t-1}) Q_{D_\phi}^{G_\theta}(Y_{1:t-1}, y_t)], \qquad (5)$$

$$\theta \leftarrow \theta + \alpha \nabla_\theta J(\theta), \qquad (6)$$

where $\alpha$ is the learning rate. SeqGAN updates the Discriminator and Generator until the stopping criteria are satisfied. An LSTM, GRU or other RNN network for the Generator and a Convolutional Neural Network (CNN) network for the Discriminator have been shown to provide good results for classification tasks [8].

The SeqGAN architecture is shown in Figure 3. The orange circles denote words in real sentences and the blue circles denote words in generated sentences. First, the Generator is pretrained with real data using the cross-entropy loss function which minimizes the negative log-likelihood. Then it is used to generate data and the Discriminator is pretrained with both generated and real data. The MC search parameters ($\beta$) are set to be the same as the Generator parameters ($\theta$). As shown on the right, an MC search is used to compute the reward for an intermediate state. This search generates $N$ complete sentences from the current state. A reward is computed for each sentence and averaged as the intermediate reward except in the last time step where the reward is obtained from the Discriminator. The input of each time step is the output of the previous time step and the next word is obtained via a multinomial distribution over the log softmax of the GRU output. Then the Generator is trained with the policy gradient. Finally, the updated Generator is used to generate data and the Discriminator is trained with both the generated and real data.

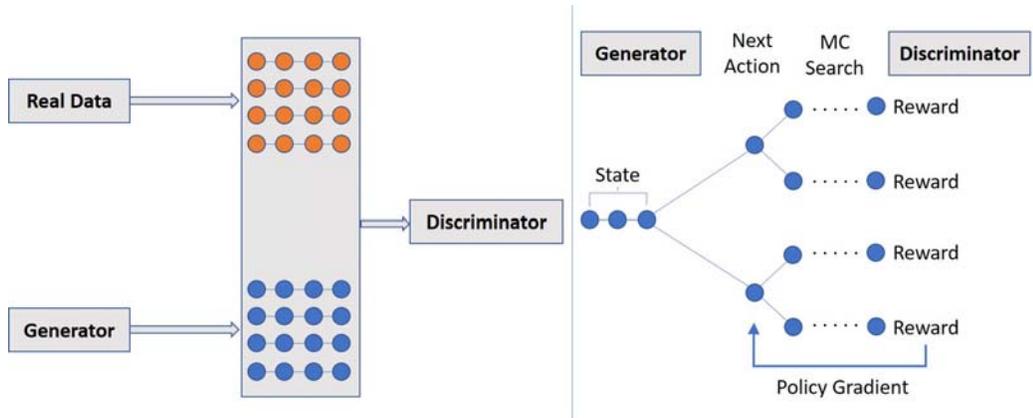

Figure 3. The SeqGAN architecture. The Discriminator on the left is trained with real data and data generated using the Generator. The Generator on the right is trained using the policy

gradient. The reward is computed by the Discriminator and this is used to determine the intermediate action values for the MC search.

### 2.4. Proposed Model

The proposed model has three steps. The first is generating log messages using SeqGAN for oversampling. The log messages are divided into two data sets, positive labeled data (normal) and negative labeled data (abnormal). Additional negative labeled data is generated using the negative labeled data set. The initial negative data set is split into sets (the Openstack data set is split into two sets and the BGL data set is split into seven sets), and fed into the SeqGAN separately. This ensures better convergence and provides different negative log messages. Further, the network speed is faster which is important with data generation. A CNN is used in the SeqGAN for the discriminator and a GRU as the generator. The GRU has one hidden layer of size 30 with the ADAM optimizer and the batch size is 128. The generated negative log messages are concatenated with the original negative data and similar messages are removed. The resulting data set is balanced with similar numbers of positive and negative data.

The second step is the Autoencoder which has two networks (positive and negative) with three hidden layers (two encoder layers and one decoder layer). The encoder layers have 400 (with L1 regularizer) and 200 neurons and the decoder layer has 200 neurons. The output layer has 40 neurons which is the same size as the input layer. The positive labeled data is fed into the positive Autoencoder. Note that this network is trained with just positive label data. The maximum number of epochs is 100 and the batch size is 128. Dropout with probability 0.8 and early stopping is used to prevent overfitting. Categorical cross-entropy loss with the ADAM optimizer is used for training. The network output is labeled as positive. The negative labeled data which has been oversampled is fed into the negative Autoencoder and the network output is labeled as negative. The two sets of labeled data are then concatenated, duplicates are removed and Gaussian noise with zero mean and variance 0.1 is added to avoid overfitting [18].

The final step is the GRU network for anomaly detection and classification. First, the concatenated data set is divided into training and testing sets with 5% for training and 95% for testing, and these sets are shuffled. The training set is then divided into two sets with 5% for training and 95% for validation. The data is fed into the GRU hidden layer of size 100 and is classified using softmax activation. 10-fold cross-validation is used in training with a maximum of 100 epochs and a batch size of 128. Dropout with probability 0.8 and early stopping is used to prevent overfitting. Categorical cross-entropy loss with the ADAM optimizer is used for training. The proposed model is shown in Figure 4.

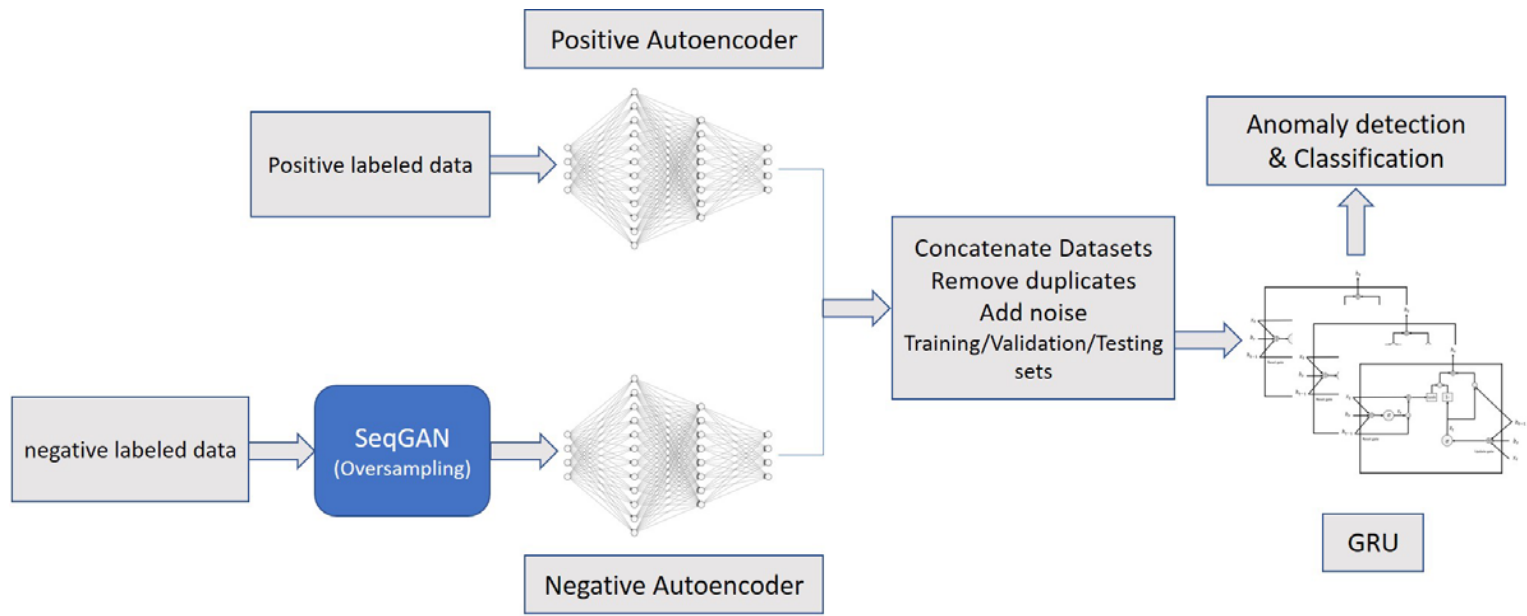

Figure 4. The proposed model architecture with SeqGAN for oversampling log messages, two Autoencoder networks and a GRU network for anomaly detection and classification.

## 3. RESULTS

In this section, the proposed model is evaluated with and without SeqGAN oversampling using the BGL and Openstack data sets. Four criteria are used to evaluate the performance, namely accuracy, precision, recall, and F-measure. Accuracy is the fraction of the input data that is correctly predicted and is given by

$$A = \frac{T_p + T_n}{T_p + T_n + F_p + F_n}, \tag{7}$$

where $T_p$ is the number of positive instances predicted by the model to be positive, $T_n$ is the number of negative instances predicted to be negative, $F_p$ is the number of negative instances predicted to be positive, and $F_n$ is the number of positive instances predicted to be negative. Precision is given by

$$P = \frac{T_p}{T_p + F_p}, \tag{8}$$

and recall is

$$R = \frac{T_p}{T_p + F_n}. \tag{9}$$

The F-measure is the harmonic mean of recall and precision which can be expressed as

$$F = \frac{2 \times P \times R}{P + R}. \tag{10}$$

All experiments were conducted on the Compute Canada Cedar cluster with 24 CPU cores, 125 GB memory and four P100 GPUs with Python in Keras and Tensorflow. We did not tune the hyperparameters of the proposed model so the default values were used for all data sets. For each data set, the average training accuracy, average validation accuracy, average training loss, testing accuracy, precision, recall, and F-measure were obtained. Tables 1 and 2 give the results for the BGL and Openstack data sets without and with SeqGAN oversampling, respectively.

### 3.1. BGL

The BlueGene/L (BGL) data set consists of 4,399,503 positive log messages and 348,460 negative log messages (without oversampling). From this data set, 11,869 logs are used for training, 225,529 for validation and the remaining 4,510,565 for testing with approximately 95% positive and 5% negative messages in each group. Without oversampling, the average training accuracy is 97.8% and average validation accuracy is 98.6% with standard deviations of 0.02 and 0.01, respectively, in 10-fold cross-validation. The average training loss is 0.07 with a standard deviation of 0.01. The testing accuracy is 99.3% with a precision of 98.9% for negative logs and 99.3% for positive logs, and recall of 91.6% and 99.9% for negative and positive logs, respectively. The F-measure is 95.1% and 99.6% for negative and positive logs, respectively.

Oversampling of the negative log messages with SeqGAN increased the number in the BGL data set to 4,137,516 so the numbers of positive and negative log messages are similar. From this data set, 21,342 logs are used for training, 405,508 for validation and the remaining 8,110,169 for testing with similar numbers of positive and negative log messages in each group.

The average training accuracy is 98.3% and average validation accuracy is 99.3% with a standard deviation of 0.01 in 10-fold cross-validation. The average training loss is 0.05 with a standard deviation of 0.01. The testing accuracy is 99.6% with a precision of 99.8% for negative logs and 99.4% for positive logs, and recall of 99.3% and 99.8% for negative and positive logs, respectively. The F-measure is 99.6% for both negative and positive logs. The accuracy levels are better than the 98% obtained with the LogSig algorithm and the BGL data set [19]. The average precision, recall and F-measure with oversampling are 99.6%, 99.5%, and 99.6%, respectively, which are better than the values of 99%, 75%, and 85%, respectively, with SVM supervised learning and 83%, 99% and 91%, respectively, with unsupervised learning [20].

### 3.2. OPENSTACK

The Openstack data set without oversampling consists of 137,074 positive log messages and 18,434 negative log messages. From this data set, 6,608 logs are used for training, 1,167 for validation and the remaining 147,733 for testing with approximately 95% positive and 5% negative messages in each group. Without oversampling, the average training accuracy is 98.4% and average validation accuracy is 97.2% with a standard deviation of 0.01 in 10-fold cross-validation. The average training loss is 0.05 with a standard deviation of 0.01. The testing accuracy is 98.3% with a precision of 97.9% for negative logs and 98.3% for positive logs, and recall of 87.1% and 99.8% for negative and positive logs, respectively. The F-measure is 92.2% and 99.0% for negative and positive logs, respectively.

Oversampling of the negative log messages with SeqGAN increased the number in the Openstack data set to 154,202. From this data set, 12,378 logs are used for training, 2,185 for validation and the remaining 276,713 for testing with similar numbers of positive and negative log messages in each group. With oversampling, the average training accuracy is 98.0% and average validation accuracy is 98.7% with a standard deviation of 0.01 in 10-fold cross-validation. The average training loss is 0.06 with a standard deviation of 0.01. The testing accuracy is 98.9% with a precision of 99.6% for negative logs and 98.2% for positive logs, and recall of 98.4% and 99.5% for the negative and positive logs, respectively. The F-measure is 99.0% and 98.8% for negative and positive logs, respectively. The accuracy levels are better than the 87.1% obtained with the IPLoM algorithm and the Openstack data set [21]. The average precision, recall and F-measure with oversampling are 98.9%, 99.0%, and 98.9%, respectively, which are better than the 94%, 99% and 97% obtained with the Deeplog network [22].

Table 1. Results without oversampling for the BGL and Openstack data sets (numbers in parenthesis are standard deviation). Positive labels are denoted by 1 and negative labels by 0.

| Data set | Average Training Accuracy | Average Validation Accuracy | Average Training Loss | Testing Accuracy | Label | Precision | Recall | F-measure |
|---|---|---|---|---|---|---|---|---|
| BGL | 97.8% | 98.6% | 0.07 | 99.3% | 0 | 98.9% | 91.6% | 95.1% |
|  | (0.02) | (0.01) | (0.01) |  | 1 | 99.3% | 99.9% | 99.6% |
| Openstack | 98.4% | 97.2% | 0.05% | 98.3% | 0 | 97.9% | 87.1% | 92.2% |
|  | (0.01) | (0.01) | (0.01) |  | 1 | 98.3% | 99.8% | 99.0% |

Table 2. Results with oversampling using SeqGAN for the BGL and Openstack data sets (numbers in parenthesis are standard deviation). Positive labels are denoted by 1 and negative labels by 0.

| Data set | Average Training Accuracy | Average Validation Accuracy | Average Training Loss | Testing Accuracy | Label | Precision | Recall | F-measure |
|---|---|---|---|---|---|---|---|---|
| BGL | 98.3% (0.01) | 99.3% (0.01) | 0.05 (0.01) | 99.6% | 0 | 99.8% | 99.3% | 99.6% |
|  |  |  |  |  | 1 | 99.4% | 99.8% | 99.6% |
| Openstack | 98.0% (0.01) | 98.7% (0.01) | 0.06% (0.01) | 98.9% | 0 | 99.6% | 98.4% | 99.0% |
|  |  |  |  |  | 1 | 98.2% | 99.5% | 98.8% |

### 3.3. DISCUSSION

The proposed oversampling with SeqGAN provided good results for both the BGL and Openstack data sets. It is evident that oversampling significantly improved the model accuracy for negative log messages. For the BGL data set, the precision, recall and F-measure after oversampling increased from 98.9% to 99.8%, 91.6% to 99.3% and 95.1% to 99.6% which are 0.9%, 7.7% and 4.5% higher, respectively. For the Openstack data set, the precision, recall and F-measure after oversampling increased from 97.9% to 99.6%, 87.1% to 98.4% and 92.2% to 99.0% which are 1.7%, 11.3% and 6.8% higher, respectively. These results show that data balancing should be considered with deep learning algorithms to improve the accuracy, especially for small numbers of minor label samples. The proposed model was evaluated with two data sets for anomaly detection and classification with only a small portion (less than 1%) used for training. This is an important result because deep learning algorithms typically require significant amounts of data for training. Note that good results were obtained even though the hyperparameters were not tuned.

The first step in the proposed model where logs are oversampled with a SeqGAN network is the most important. These networks have been shown to provide promising results in generating text such as poems [23]. The concept of generating data is similar to that for oversampling. It was surprising that duplication in the oversampled log data was not high (less than 5%). As a consequence, after removing duplicates there was a significant amount of data available for anomaly detection and classification using deep learning. The second step which extracts features from the data using an Autoencoder is also important. The Autoencoder output is very suitable for use with an RNN based algorithm such as a GRU for anomaly detection and classification. The results obtained show that the proposed model can provide excellent results even when the data is imbalanced.

### 4. CONCLUSIONS

In this paper, a model was proposed to address the problem of imbalanced log messages. In the first step, the negative logs were oversampled with a SeqGAN network so that the numbers of positive and negative logs are similar. The resulting labeled logs were then fed into an Autoencoder to extract features and information from the text data. Finally, a GRU network was used for anomaly detection and classification. The proposed model was evaluated using two log message data sets, namely BGL and Openstack. Results were presented which show that oversampling can improve detection and classification accuracy. In the future, other text-based GAN networks such as TextGAN and MaliGAN can be used for oversampling.

### REFERENCES

[1] T. Munkhdalai, O.-E. Namsrai and K. H. Ryu, "Self-training in Significance Space of Support Vectors for Imbalanced Biomedical Event Data", *BMC Bioinformatics*, vol. 16, no. S7, pp. 1-8, 2015.

[2] Y.-H. Liu and Y.-T. Chen, "Total Margin Based Adaptive Fuzzy Support Vector Machines for Multiview Face Recognition", in *IEEE International Conference on Systems, Man and Cybernetics*, pp. 1704–1711, 2005.

[3] M. J. Siers and M. Z. Islam, "Software Defect Prediction using A Cost Sensitive Decision Forest and Voting, and A Potential Solution to the Class Imbalance Problem", *Information Systems*, vol. 51, pp. 62-71, 2015.

[4] N. V. Chawla, "Data Mining for Imbalanced Datasets: An Overview", in *Data Mining and Knowledge*

*Discovery Handbook*, pp. 853–867, Springer, 2005.

[5] I. Goodfellow, J. Pouget-Abadie, M. Mirza, B. Xu, D. Warde-Farley, S. Ozair, A. Courville and Y. Bengio, "Generative Adversarial Nets", in *International Conference on Neural Information Processing Systems*, pp. 2672–2680, MIT Press, 2014.

[6] D. Li, Q. Huang, X. He, L. Zhang and M.-T. Sun, "Generating Diverse and Accurate Visual Captions by Comparative Adversarial Learning", *arXiv e-prints,* arXiv:1804.00861, 2018.

[7] L. Liu, Y. Lu, M. Yang, Q. Qu, J. Zhu and H. Li, "Generative Adversarial Network for Abstractive Text Summarization", *arXiv e-prints,* p. arXiv:1711.09357, 2017.

[8] L. Yu, W. Zhang, J. Wang and Y. Yu, "SeqGAN: Sequence Generative Adversarial Nets with Policy Gradient", in *AAAI Conference on Artificial Intelligence*, pp. 2852–2858, 2017.

[9] D. E. Rumelhart, G. E. Hinton and R. J. Williams, "Parallel Distributed Processing: Explorations in the Microstructure of Cognition", in *Parallel Distributed Processing: Explorations in the Microstructure of Cognition,* Vol. 1, D. E. Rumelhart, J. L. McClelland and C. PDP Research Group, Eds., pp. 318–362, MIT Press, 1986.

[10] K. Cho, B. Merriënboer, C. Gulcehre, D. Bahdanau, F. Bougares, H. Schwenk and Y. Bengio, "Learning Phrase Representations Using RNN Encoder–Decoder for Statistical Machine Translation", in *Conference on Empirical Methods in Natural Language Processing*, pp. 1724–1734, 2014.

[11] D. J. Rezende, S. Mohamed and D. Wierstra, "Stochastic Backpropagation and Approximate Inference in Deep Generative Models", in *International Conference on Machine Learning*, vol. 32, pp. II–1278–II–1286, 2014.

[12] I. Higgins, L. Matthey, A. Pal, C. Burgess, X. Glorot, M. Botvinick, S. Mohamed and A. Lerchner, "Beta-VAE: Learning Basic Visual Concepts with a Constrained Variational Framework", in *International Conference on Learning Representations*, 2017.

[13] M. Kuta, M. Morawiec and J. Kitowski, "Sentiment Analysis with Tree-Structured Gated Recurrent Units", in *Text, Speech, and Dialogue*, Lecture Notes in Computer Science, pp. 74–82, Springer, 2017.

[14] K. Irie, Z. Tüske, T. Alkhouli, R. Schlüter and H. Ney, "LSTM, GRU, Highway and a Bit of Attention: An Empirical Overview for Language Modeling in Speech Recognition", in *Interspeech*, pp. 3519–3523, 2016.

[15] Y. LeCun, Y. Bengio, and G. Hinton, "Deep Learning", *Nature*, vol. 521, no. 7553, pp. 436–444, 2015.

[16] S. Hochreiter and J. Schmidhuber, "Long Short-Term Memory", *Neural Comput.,* vol. 9, no. 8, pp. 1735-1780, 1997.

[17] R. S. Sutton, D. McAllester, S. Singh and Y. Mansour, "Policy Gradient Methods for Reinforcement Learning with Function Approximation", in *International Conference on Neural Information Processing Systems*, pp. 1057–1063, MIT Press, 1999.

[18] H. Noh, T. You, J. Mun and B. Han, "Regularizing Deep Neural Networks by Noise: Its Interpretation and Optimization", in *Advances in Neural Information Processing Systems*, (I. Guyon, U. V. Luxburg, S. Bengio, H. Wallach, R. Fergus, S. Vishwanathan, and R. Garnett, eds.), vol. 30, pp. 5109–5118, Curran Associates, 2017.

[19] P. He, J. Zhu, S. He, J. Li and M. R. Lyu, "An Evaluation Study on Log Parsing and Its Use in Log Mining", in *IEEE/IFIP International Conference on Dependable Systems and Networks*, pp. 654–661,

2016.

[20] S. He, J. Zhu, P. He and M. R. Lyu, "Experience Report: System Log Analysis for Anomaly Detection", in *IEEE International Symposium on Software Reliability Engineering*, pp. 207–218, 2016.

[21] J. Zhu, S. He, J. Liu, P. He, Q. Xie, Z. Zheng and M. R. Lyu, "Tools and Benchmarks for Automated Log Parsing", in *International Conference on Software Engineering: Software Engineering in Practice*, *International Conference on Software Engineering: Software Engineering in Practice*, pp. 121–130, 2019.

[22] M. Du, F. Li, G. Zheng and V. Srikumar, "DeepLog: Anomaly Detection and Diagnosis from System Logs through Deep Learning", in *ACM Conference on Computer and Communications Security*, pp. 1285–1298, 2017.

[23] X. Wu, M. Klyen, K. Ito and Z. Chen, "Haiku Generation using Deep Neural Networks", *The Association for Natural Language Processing*, pp. 1–4, 2017.